%% file: main.tex
\documentclass[letterpaper, 10 pt, conference]{ieeeconf}  

\IEEEoverridecommandlockouts                              
\usepackage{times} 
\usepackage{amsmath} 
\usepackage{amssymb}  

\title{Prospection: Interpretable Plans From Language By Predicting the Future}

\author{Chris Paxton$^{1}$, Yonatan Bisk$^{2}$, Jesse Thomason$^{2}$, Arunkumar Byravan$^{2}$, Dieter Fox$^{1,2}$
\thanks{$^{1}$NVIDIA, USA
        }
\thanks{$^{2}$Paul G. Allen School of Computer Science and Engineering, University of Washington, Seattle, USA}
}

\usepackage[utf8]{inputenc}
\usepackage{hyperref}
\usepackage{booktabs}
\usepackage{multirow}
\usepackage{subcaption} 
\usepackage{array}
\usepackage{xspace}
\usepackage{flushend}
\usepackage{algorithm}
\usepackage{algorithmic}
\usepackage{makecell}

\newcommand{\beginsupplement}{%
        \setcounter{table}{0}
        \renewcommand{\thetable}{A\arabic{table}}%
        \setcounter{figure}{0}
        \renewcommand{\thefigure}{A\arabic{figure}}%
        \setcounter{section}{0}
        \renewcommand{\thesection}{A-\roman{section}}%
     }

\hypersetup{urlcolor=blue, colorlinks=true}

\usepackage[colorinlistoftodos,prependcaption,textsize=tiny]{todonotes}

\newcommand{\dreamcell}{\textsc{DreamCell}\xspace}
\newcommand{\dreamcells}{\textsc{DreamCells}\xspace}

\newcommand{\arxiv}[2]{#1}

\begin{document}

\maketitle

\begin{abstract}
High-level human instructions often correspond to behaviors with multiple implicit steps.
In order for robots to be useful in the real world, they must be able to to reason over both motions and intermediate goals implied by human instructions.
In this work, we propose a framework for learning representations that convert from a natural-language command to a sequence of intermediate goals for execution on a robot.
A key feature of this framework is \emph{prospection}, training an agent not just to correctly execute the prescribed command, but to predict a horizon of consequences of an action before taking it.
We demonstrate the fidelity of plans generated by our framework when interpreting real, crowd-sourced natural language commands for a robot in simulated scenes.
\end{abstract}


\section{Introduction}
\label{sec:introduction}
\input{01_introduction.tex}

\begin{figure*}[t]
\centering
\includegraphics[width=2\columnwidth]{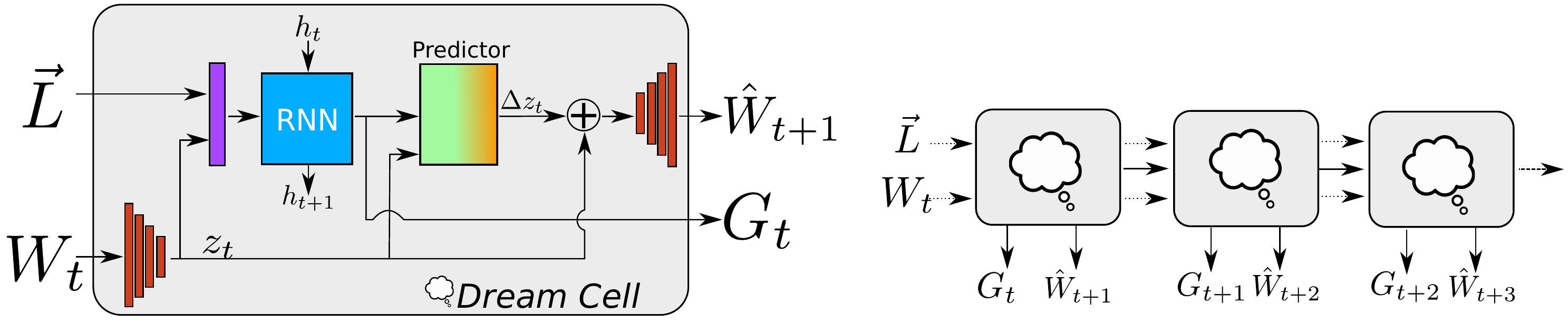}
\caption{
At each timestep, the model receives an LSTM encoded language vector $\vec{L}$, the initial world state $W_0$, and the current world state $W_t$.
Using these, it predicts the next world state $\hat{W}_{t+1}$ and sub-goal $G_t$.
}
\label{fig:system}
\end{figure*}

\section{Related Work}
\label{sec:relatedwork}
\input{02_relatedwork.tex}

\section{Problem Description}
\label{sec:problem}
\input{03_problem.tex}

\section{Approach}
\label{sec:approach}
\input{04_approach.tex}

\section{Experiment Setup}
\label{sec:experiment}
\input{05_experiment.tex}

\section{Results}
\label{sec:results}
\input{06_results.tex}

\section{Conclusions}
\label{sec:conclusions}
\input{07_conclusions.tex}

\section{Acknowledgements}

This work was funded in part by the National Science Foundation under contract no. NSF-NRI-1637479, and the DARPA CwC program through ARO
(W911NF-15-1-0543). We would like to thank Jonathan Tremblay for valuable discussions.

\bibliographystyle{IEEEtran}
\bibliography{main}

\arxiv{
\clearpage
\begin{onecolumn}
\input{08_Appendix.tex}
\end{onecolumn}
}{}

\end{document}

%% file: 01_introduction.tex
A robot agent executing natural language commands must solve a series of problems.
First, human language must be translated to an understanding of intent.
For example, the command \emph{pick up the yellow block and place it on top of the red block} corresponds to an intended change in world state that results in a yellow block on top of a red one.
Given that understanding, an agent must plan a sequence of actions it can take to reach the target world state.
In the above example, this could be \emph{(move(yellow); grasp(yellow); move(yellow, red); release(yellow))}.
Finally, these high level controls have to be executed in the world by servoing an arm to appropriate positions and controlling the gripper.

Each of these problems is challenging and has been investigated by existing research.
Commonly, a pipeline approach is used, where each problem is addressed sequentially, and the outputs of one are fed to the next.
In the example above, the semantic understanding that the goal is \emph{on(yellow, red)} from natural language is passed to a high level controller.
To simplify high level control prediction, robot perception is often augmented with visual semantics information, such as oracle object detections, bounding boxes, or 6d poses~\cite{paxton2017costar,xu2018neural}.
In this work, we instead train a single model end-to-end that takes natural language and raw pixels, depths, and joint states to produce low-level controls to accomplish a goal.

Additionally, rather than treat the pipeline problems above independently, we introduce a \emph{prospection} component to an agent's training and inference, which facilitates ``dreaming'' about the consequences of chosen high level actions in the current scene.
Prospection is the ability to reason about the consequences of future actions without executing them~\cite{gilbert:science07}.
Our approach allows the agent to predict whether its high level actions will lead to undesirable world states.

\arxiv{
\begin{figure}[t]
\centering
\includegraphics[width=\columnwidth]{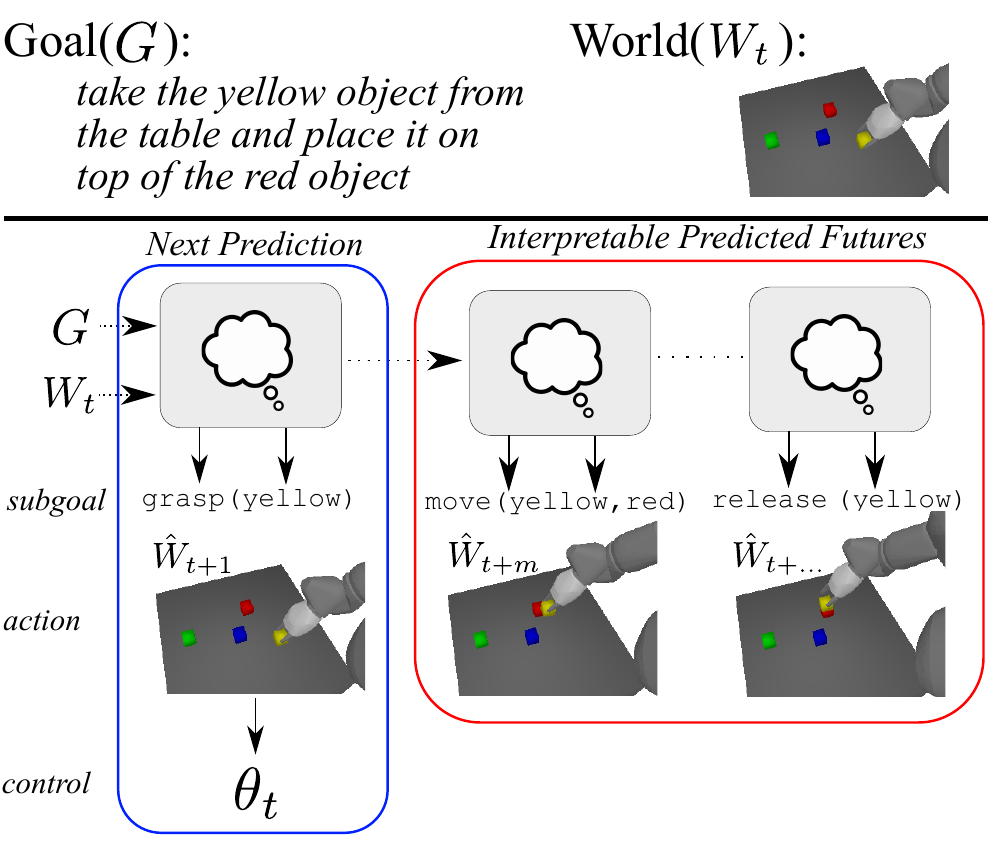}
\caption{\dreamcells convert instructions to interpretable subgoals which can be visualized ($\hat{W}$) and executed ($\theta$).}
\label{fig:intro}
\end{figure}
}{
\begin{figure}[t]
\centering
\includegraphics[width=\columnwidth]{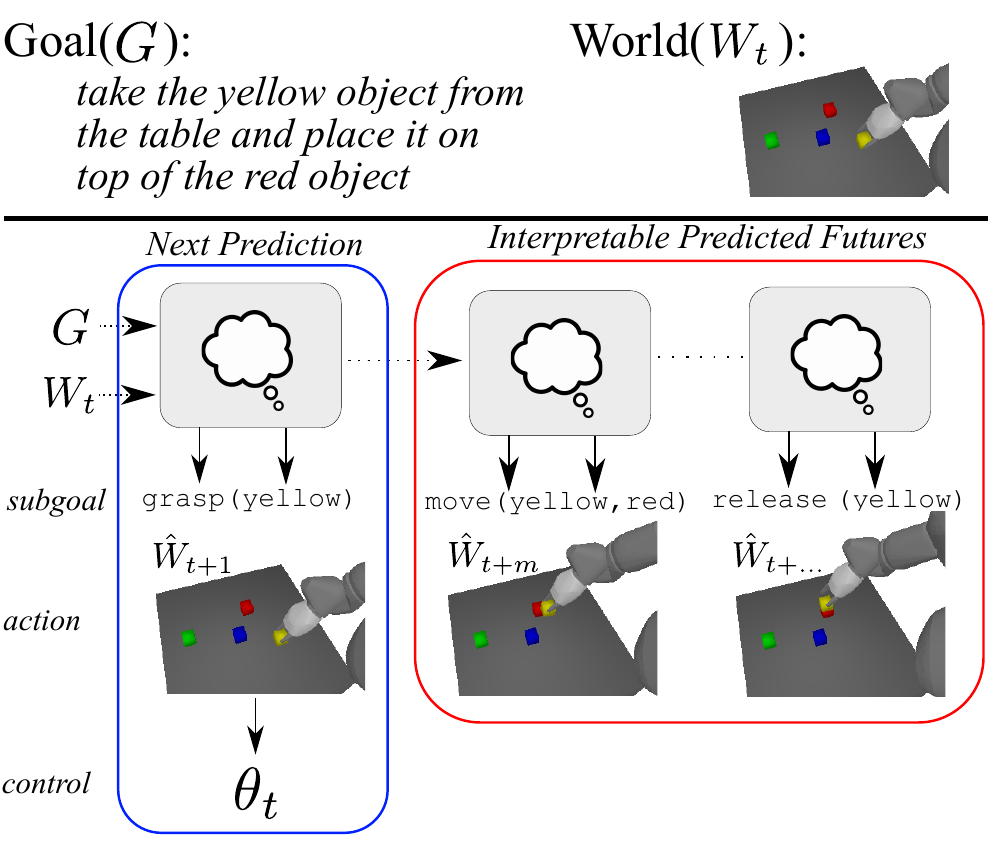}
\caption{\dreamcells convert instructions to interpretable subgoals which can be visualized ($\hat{W}$) and executed ($\theta$).}
\label{fig:intro}
\vskip -0.25cm
\end{figure}
}

We consider a simple pick-and-place task, where the goal is to stack one block on top of another (Figure~\ref{fig:intro}).
This is limited in that there are only a few high-level actions the robot can take in a given world.
Still, it proves challenging when specified with real, crowd-sourced natural language.

In short, our contributions are:
\begin{itemize}
    \item A language embedding rich enough to specify a sequence of actions to achieve a task-level goal.
    \item A \dreamcell architecture to predict future world states from language and raw sensor observations.
    \item An approach to convert a task plan output from these \dreamcells into low-level executions.
\end{itemize}

%% file: 02_relatedwork.tex
Our work draws inspiration from recent efforts on learning abstract representations for planning~\cite{paxton2018visual}.
We build primarily on work in planning and natural language processing, with important future work in manipulation.

Communicating control and goals has traditionally been accomplished by specifying
high level operations~\cite{intera5,paxton2017costar,paxton2018evaluating},  via formal languages like 
the Problem Domain Description Language (PDDL)~\cite{ghallab1998pddl} or as a Hierarchical Task Network~\cite{erol1994htn}.
Such systems provide a straightforward way to compose black-box operations to solve problems.  While we maintain an interpretable intermediate layer, our interface 
is natural language, most akin to~\cite{Paul:2016, Arumugam:2017} though we work in a
fully end-to-end differentiable paradigm where embedded language representations are
learned alongside visual encodings of the world.

Our work is also motivated by the Universal Planning Networks which learn an embedding for images and a world state vector used to generate motion plans to goals specified via a target image~\cite{srinivas2018universal}.
That work learns a distance metric from the current state image to the target image which is used to perform rollouts for training and inference. 
While learned generic representations have notions of agency and planning, the produced plans lack human interpretability, which may be important to modularity and generalization~\cite{garnelo16:drl}.
Neural approaches and scaling robotic learning within simulation have become common
as they allow for end-to-end training and can easily acquire more 
data (often from multiple domains) than otherwise possible on a physical device~\cite{xu2018neural}.  This has proven particularly important for RL-based approaches \cite{nachum2018dataefficient} and interpretability \cite{tremblay2018synth}.
More generally there is a growing literature on learning deep representations that can be used to accomplish local control tasks or simple object manipulations~\cite{byravan2017se3,finn2017deep,srinivas2018universal,weber2017imagination}.

Core to our contribution is simulating the future actions and dynamics of our
system.  High level process simulation has been used in the NLP literature~\cite{bosselut2018simulating} without sensor data.
Simultaneously, prospection has been used before in RL, often as a means of model-based control~\cite{lenz2015deepmpc,weber2017imagination,pascanu2017learning}, and for fine control tasks like cutting~\cite{lenz2015deepmpc}. In addition, our approach is compatible with work in Visual Robot Task Planning~\cite{paxton2018visual}, which shows that prospective subgoal predictions can be used to generate task plans.

Complementary to our work is the growing literature in NLP focusing on complex grounding instructions with esoteric 
references and other long tail linguistic phenomena~\cite{bisk2016natural,bisk2017learning}. Natural language communication 
with robotics also allows for learning joint multimodal representations~\cite{thomason:ijcai16,thomason2018guiding} which harness the unique perceptual and manipulation capabilities of robotics.

While accurate grasping and placement is not a focus of our work, it has been explored in the literature~\cite{levine2016learning, mahler2017dex,morrison2018closing}. In particular, high-precision grasping with deep neural networks generally takes the form of predicting a grasp success classifier~\cite{levine2016learning, mahler2017dex}.

%% file: 03_problem.tex
Given a natural language command and raw sensor readings for a scene, the task is to issue a sequence of low level controls to a robotic arm that accomplish the intended task. 

{\bf Specification.}
A \dreamcell takes in the initial $W_0$ and current $W_t$ simulation-based state observations and a natural-language sentence $s$ to produce a sequence of intermediate, latent-space goals $z_i, \dots, z_{i+h}$ for this pick-and-place task up to horizon $h$.
Each goal is a semantically meaningful break point in the execution, e.g., a completed grasp on the target.

\dreamcells make three predictions at every time step:
\begin{enumerate}
    \item A sequence of subgoals predictions, representing the next high level actions out to some planning horizon;
    \item A sequence of hidden state representations $z_i, \dots, z_{i+h}$ representing the results of these subgoals; and
    \item The end effector command $\theta$ that parameterizes the low-level controller for task execution, consisting of a 6DOF pose and gripper command.
\end{enumerate}

These predictions allow us to learn an interpretable, executable representation for hallucinating future world states.

{\bf Metrics.}
We measure both \emph{extrinsic} performance on the task and \emph{intrinsic} performance of \dreamcell components.
Extrinsically, we evaluate how well the robot agent completes the pick-and-place task.
We record binary task success/failure as whether the target block to be moved is dropped within a threshold of its intended position based on the language command.
We also record the average mean-squared error of the predicted end effector goal at each step of the task.
This metric penalizes moving the wrong block while giving partial credit for moving the right block to towards the right place, even if it never arrives there or is placed unstably (e.g., if it falls off the target block after \emph{release}).
Intrinsically, we evaluate the language-to-action component by how closely the predicted sequence of subgoals matches ground truth execution.

%% file: 04_approach.tex
\arxiv{
\begin{figure*}[t]
\centering
\includegraphics[width=1.8\columnwidth]{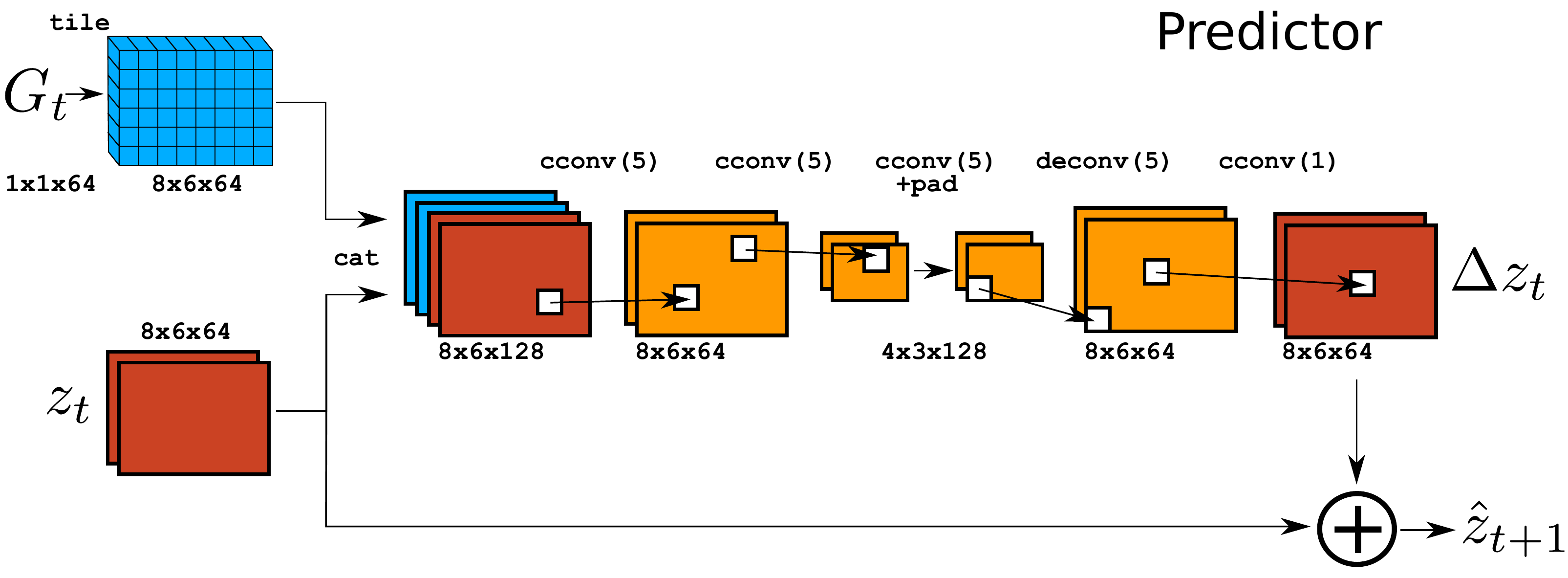}
\caption{
Diagram of a single prediction cell. The prediction cell predicts a change in hidden state $\Delta z$, and is an important component of the \dreamcell, used both for visualizing possible futures and for predicting the goal of a particular motion for execution on our robot.}
\label{fig:prediction*}
\end{figure*}
}{
\begin{figure}[t]
\includegraphics[width=\columnwidth]{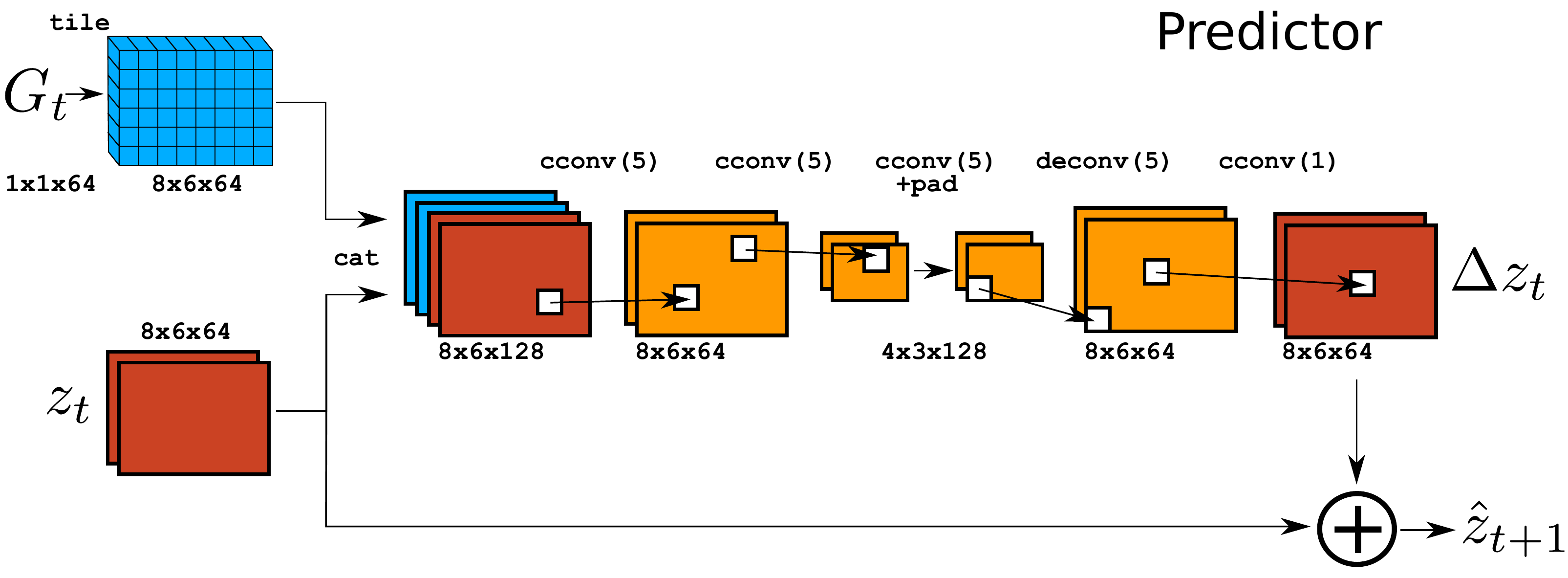}
\caption{
Diagram of a single prediction cell. The prediction cell predicts a change in hidden state $\Delta z$, and is an important component of the \dreamcell.}
\label{fig:prediction*}
\vskip -0.5cm
\end{figure}
}

We train the system end-to-end using simulated data.
This allows us to automatically generate training sequences for both images and
high-level subgoals.  Subgoals take the form of semantic predicates like
{\tt grasp} and {\tt move} with block arguments.
To simplify notation, throughout the paper, 
we refer to the union of 
RGB, pose, and depth images with the single world state variable $W$.
At every timestep the model is provided the current world observation ($W_t$), 
a description of the goal configuration in natural language (encoded as 
$\vec{L}$). In practice $W_t$ also includes the initial state $W_0$ to capture changes over time.
All aspects of the model (including the
encoders for both language and the world) are trained together.
The model is trained on supervised demonstration data collected from an expert policy as per previous work~\cite{byravan2017se3,paxton2018visual}.

The basic unit in our model is the \dreamcell (Fig.~\ref{fig:system}) which produces a sub-goal and a corresponding predicted image of the arm's position at the next time step.  This formulation allows 
for recurrent chaining of cells to rollout future goals and states (Fig.~\ref{fig:intro}).  
Specifically, because the output of the \dreamcell includes a deconvolved hallucination of
the next world state ($\hat{W}_{t+1}$) we can simply continue to run the network forward, where
true observations are replaced with the network's predictions.
Our cell has two outputs at every timestep $t$: 1. Subgoals ($G_t$) and 2. Predicted Worlds ($\hat{W}_{t+1}$).
We provide intrinsic evaluations on 
future prediction performance in Section \ref{sec:results}.

At inference time, we generate a task plan by rolling out multiple \dreamcell timesteps into a possible future given state observations, $z$.
The core of our approach is that these subgoals are converted into estimated world states $\hat{W}$ and associated end effector goals $\theta$, which are fed into a lower level controller $\pi$ that will convert them into trajectories.

\subsection{\dreamcell Subgoal Module}

The subgoal module predicts the next subgoal from the current world state and the language instruction.
It is formulated similarly to image captioning and sequence to sequence prediction.
First, we use an LSTM~\cite{hochreiter1997long} to encode the goal as expressed in language.  Words are embedded as 64 dimensional
vectors initialized randomly.  We concatenate the final hidden state ($\vec{L}$) with the output of our world 
encoder ($z_t$) as the initial hidden state of a new LSTM cell for decoding.

We generate an output of $G_t=({\tt verb}, {\tt to\_obj}, {\tt with\_obj})$
tuples at each timestep, to a horizon of length five.
We divide the subgoal $G_t$ into: ${\tt verb}$ the action to be taken, ${\tt to\_obj}$ the object to servo towards, and ${\tt with\_obj}$ the optional object in hand (e.g., ${\tt move(red, yellow)}$ moves the yellow block in hand to the target red block).

During training, we use cross entropy loss on all 5x3 generations.  
The first timestep in the RNN is passed the current hidden state and the 
At each timestep the RNN cell is passed the prediction from the previous timestep.
As is common practice 
in the language modeling literature \cite{press-wolf:2017:EACLshort}, we tie the emission and embedding matrix parameters.
Traditionally, 
this is achieved by simply transposing a single matrix.
Our model produces tuples by passing the hidden vector
through three different feed-forward layers, so, to re-embed predictions we multiply by the three transposed embedding
matrices and average the outputs to reconstruct an embedding.

\arxiv{
\begin{figure*}[bt]
\centering
\includegraphics[width=1.6\columnwidth]{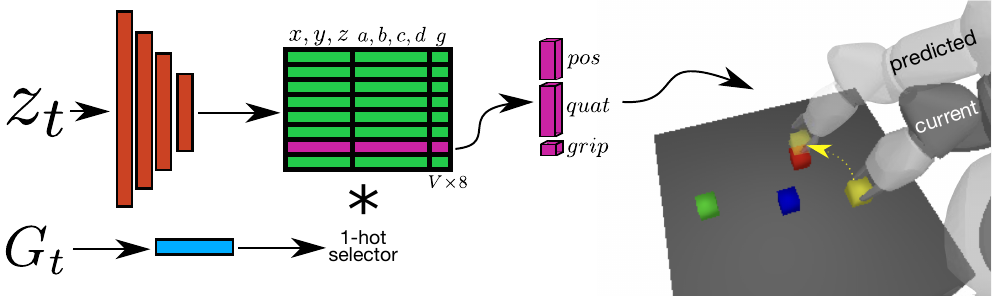}
\caption{The actor module takes in a hidden state and associated subgoal API and converts this to a motion goal, which is represented as a Cartesian $(x,y,z)$ position, a unit quaternion $q=(a,b,c,d)$, and a gripper command $g \in (0,1)$. This motion goal can then be sent to the control module for execution.}
\label{fig:actor}
\end{figure*}
}{
\begin{figure}[bt]
\includegraphics[width=\columnwidth]{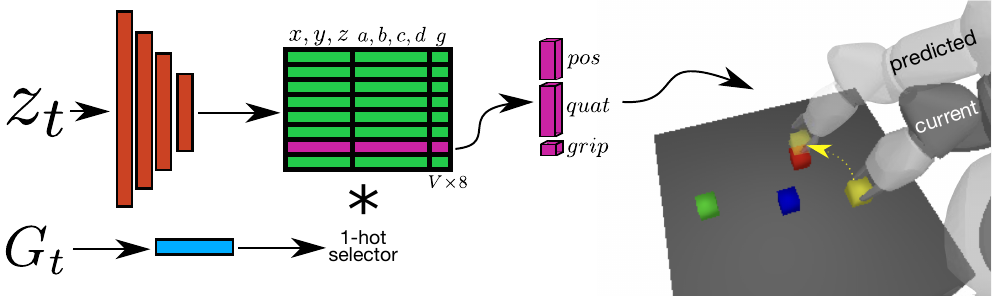}
\caption{The actor module takes in a hidden state and associated subgoal API and converts this to a motion goal. This motion goal can then be sent to the control module for execution.}
\label{fig:actor}
\vskip -0.5cm
\end{figure}
}

\subsection{\dreamcell World Predictor Module}

The prediction cell, shown in Fig.~\ref{fig:prediction*}, takes in the current hidden state $z_t$ and the predicted subgoal $G_t$.
Each prediction model outputs a predicted latent-state subgoal $\hat{z}_{t+1}$, such that $P(z_t, G_t) = \hat{z}_{t+1}$.
In effect, we learn a many-to-one mapping across multiple timesteps, all of which need to produce the same goal.
We should also be able to roll this simulation forward in time in order to visualize future actions.
Training this prediction space is a difficult problem and requires a complex loss function involving multiple components.

The prediction cell is a simple autoencoder mapping inputs $W$ to and from a learned latent space, as show in Fig.~\ref{fig:system}. 
World observations $W_t$ and $W_0$ are combined into a single estimated latent state $z_t$.
The vector containing the predicted subgoal $G_t$ is tiled onto this state. We use a bottleneck within each prediction cell to force information to propagate across the entire predicted image, and then estimate a change in latent state $\Delta z$ such that $\hat{z}_{t+1} = z_t + \Delta z$.

When visualizing the predicted image $\hat{W}_{t+1}$, we use a decoder consisting of a series of 5x5 convolutions and bilinear interpolation for upsampling.

\subsection{Actor Module}

The actor module, shown in Fig.~\ref{fig:actor}, predicts the parameters of an action that can be executed on the robot. Specifically, it takes in $z_t$ and the current high-level action and predicts a destination end effector pose that corresponds to the robot's position at $z_t$.

\arxiv{
The architecture is a simple set of convolutions: the high-level action is concatenated with the current $z_t$ as in the prediction module, then a set of three 3x3 convolutions with 64, 128, and 256 filters, each followed by a 2x2 max pool. This is followed by a dropout and a single 512 dimensional fully connected layer, and then to $N_{verbs} \times 8$ outputs, predicting gripper command $g \in (0, 1)$,
Cartesian end effector position, and a unit quaternion for each high-level action verb. The gripper command uses a sigmoid activation where $0$ represents closed and $1$ represents open, and Cartesian end effector position uses a tanh activation function. All pose values are normalized to be in $(-1, 1)$.

A one-hot attention over action verbs chooses which pose and gripper command should be executed. In effect, the actor learns to compute a set of pose features for predicting the next manipulation goal and learns a simple perceptron model for each action verb in order to choose where the arm should go and whether the gripper should be opened or closed after the motion is complete.
}{
The architecture is a simple set of convolutions: the high-level action is concatenated with the current $z_t$ as in the prediction module, then a set of three 3x3 convolutions with 64, 128, and 256 filters, each followed by a 2x2 max pool. This is followed by a dropout and a single 512 dimensional fully connected layer, and then to $N_{verbs} \times 8$ outputs, predicting gripper state,
Cartesian gripper position, and a unit quaternion for each high-level action verb. A one-hot attention over action verbs chooses which pose and gripper command should be executed. In effect, the actor learns to compute a set of pose features for predicting the next manipulation goal and learns a simple perceptron model for each verb.
}

\subsection{Training}

We train the encoder and decoder jointly when training the Prediction and Actor modules and optimize with Adam~\cite{Kingma:2015}, using an initial learning rate of $1e-3$.
We fix the latent state encoder and decoder functions after this step, then use the learned hidden space to train the Subgoal module.

\textbf{Image Reconstruction Loss} This determines how well our model can reconstruct an image from a given hidden state $z_t$, and is trained on the output of our visualization module. We used an L2 loss on pixels both for RGB and depth. Depth values were capped at 2 meters and were normalized to be between 0 and 1.

\textbf{Subgoal Recovery Loss}
Image reconstruction losses are often insufficient for capturing fine details. This issue has motivated recent work on GANs~\cite{goodfellow2014generative}. These are often unstable, so we propose an alternative solution specialized to our problem. Since each successful high-level action has a predictable result, we jointly train a classifier that will recover the subgoal associated with each successive high-level action $\hat{G}_t$. We use $C_G(z_t)$ as the classifier loss, minimizing cross entropy between the recovered estimate $\hat{G}_t$ and ground truth $G_t$.

\textbf{Actor Loss}
Instead of estimating the full joint state of the robot as the result of a high-level action, our Actor module estimates the end-effector pose $\theta_t$ associated with sugboal $G_t$.

These poses are represented as $\theta = (\hat{p}, \hat{q}, \hat{g})$, where $\hat{p}$ is the Cartesian position, $\hat{q}$ is a quaternion describing the orientation, and $\hat{g} \in (0, 1)$ is the gripper command.
When regressing to poses, we use a mixture of the L2 loss between Cartesian position and a loss derived from the quaternion angular distance (to capture both spatial and rotational error). The angle between two quaternions $\hat{q}_1$ and $\hat{q}_2$ is given as:
\[
    \omega = \cos^{-1}(2\langle \hat{q}_1, \hat{q}_2 \rangle^2).
\]
To avoid computing the inverse cosine as a part of the loss, we use a squared distance metric.
In addition, normalize gripper commands to be between 0 and 1, where 0 is closed and 1 is open, and trained with an additional L2 loss on predicted gripper commands.
Given estimated pose $\hat{\theta}$ and final pose $\theta$, we calculate pose estimation loss:
\[
    C_{actor}(\hat{\theta}, \theta) = \lambda_{actor} \left\{ \|\hat{p} - p\|_2^2 + (1 - \langle \hat{q}, q \rangle) + \|\hat{g} - g\|_2^2 \right\}.
\]

\textbf{Object Pose Estimation Loss}
It is important to ensure that our learned latent states $z_t$ capture all the necessary information to perform the task. As such, we use an augmented loss $C_{obj}(z_t)$ that predicts the position 
of each of the four blocks in the scene at the observed frame.
This information is not used at test time, but is structurally identical to the pose estimation loss $C_{pose}$

\textbf{Combined Prediction Loss}
The final loss function for predicting the effects of performing a sequence of high-level actions is then:
\begin{align*}
    C(\hat{Z}) =& \lambda_{W} \|\hat{W}_{t} - W_t\|_2^2 + C_{obj}(z_t) +\\
    & \sum_{i \in h} 
    \left( 
      {\lambda_{W} \|\hat{W}_{t+i} - W_{t+i}\|_2^2  \atop +\  C_{actor}(\hat{\theta}_{t+i}, \theta_{t+i})  + C_G(\hat{G}_{t+i}, G_{t+i})} 
     \right).
\end{align*}

\subsection{Execution}

When executing in a new environment, the robot agent takes in the current world state $W_0$ and a natural language instruction $L$.
The agent computes a future prediction using a \dreamcell, by rolling out predicted goals $G_t,\dots,G_{t+H}$ which generate latent space subgoals.
These subgoals can then be visualized to provide insight into how the robot expects the task to progress.  This also illuminates misunderstandings and limitations of the system (see Analysis \ref{sec:results}).

The system generates new prospective plans out to a given planning horizon. After predicting the next subgoal $z_{t+1}$, it will then use the actor to estimate the next motion goal $\theta_{t+1}$. This goal is sent to the low-level execution system, which in our case is a traditional motion planner that does not have knowledge of object positions.
\arxiv{In our case, the planner used was RRT-connect~\cite{kuffner2000rrt}, via MoveIt}{}

\arxiv{
In the future, these subgoals shown to the user, who can give the final confirmation on whether or not to execute this hallucinated task plan if it accomplishes what they requested.
Alternatively, the user could input a new $L$, or the agent could sample a new sequence of goals.
}{}

%% file: 05_experiment.tex
\begin{figure}[bt!]
 \centering
 \begin{tabular}{@{}cc@{}}
  \frame{\includegraphics[trim={2cm 0 0 2.8cm},clip,width=0.4\linewidth]{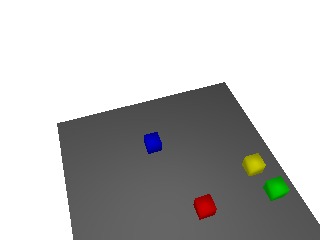}} 
  &  \frame{\includegraphics[trim={2cm 0 0 2.8cm},clip,width=0.4\linewidth]{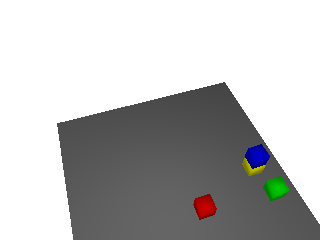}}\\
  \multicolumn{2}{@{}c@{}}{\emph{put the blue cube onto the yellow cube}} \\
  \multicolumn{2}{@{}c@{}}{\emph{stack the top most cube onto the second highest cube}} 
 \end{tabular}
\caption{Human participants on Mechanical Turk gave two commands for how to create the target image (right) from the initial image (left).
}
\label{fig:nlp_mturk}
\vskip -0.3cm
\end{figure}

We performed a number of variations on a simple block stacking task. All experiments were performed in simulation. We collected 5015 trials using a sub-optimal expert policy, of which 2370 were successes. Our model was trained only on successful examples.

We generated trials using a simple simulation of an ABB YuMi robot picking up 3.5~cm cubes.
There were four cubes, one of each color: red, green, yellow, blue.
When collecting data, we first randomly compute a table position within a 50~cm box centered in front of the robot. Blocks were randomly placed in non-intersecting positions on this table. The arm's initial position was also randomized to an area off the right side of the table.

We selected manipulation goals at random, and provided a simple expert policy which moved to pick up each object using an RRT motion planner. The plan has five steps: \texttt{align} with the top of a random block, \texttt{grasp} that block and close the gripper, \texttt{lift} the block off the table, \texttt{move} the block to atop another block, and then \texttt{release} the block.

We collected natural language commands from human annotators through the Mechanical Turk crowd-sourcing platform.\footnote{\url{https://www.mturk.com/}}
Annotators were shown two scene images: one before and one after a block had been stacked on another block.
They were instructed to give two distinct commands that would let someone create the second scene from the first (Figure~\ref{fig:nlp_mturk}), and were paid \$0.25 per such annotation.
For each of our 2370 successful trials, we obtained two language commands describing the high level pick-and-place goal.
On average, commands are 11 words long.\footnote{Participants used 389 unique words after lowercasing and tokenization.} We compare Turk data to unambiguous templated language that was procedurally generated from the manipulated blocks.

%% file: 06_results.tex
\begin{table}[bt]
    \centering
    \begin{tabular}{l c c c c c c}
        \toprule
        & \multicolumn{5}{c}{L2 distance in cm $\downarrow$} & Success $\uparrow$\\
        & Align & Grasp & Lift & Move & Release & Rate \\
        \midrule
        Oracle & 0.04 & 0.03 & 0.04 & 0.04 & 0.04 & 98.4\% \\
        GT Action   & 0.32 & 0.31 & 0.48 & 0.63 & 0.63 & 90.4\%\\
        \midrule
        Template & 0.32 & 0.39 & 0.47 & 0.65 & 0.65 & 87.8\%\\
        Real Lang & 0.51 & 1.23 & 1.50 & 2.39 &  2.40 & 77.1\% \\
        \bottomrule
    \end{tabular}
    \caption{L2 distances and accuracy when executing plans generated from either ambiguous natural language instructions or unambiguous template language.}
    \label{table:extrinsic-ee-pose}
    \vskip -0.5cm
\end{table}

 We ran a set of experiments on our simulation, and computed task execution success rate. We analyze the performance of the Subgoal and Predictor modules given different classes of language input.
 
\begin{figure*}[t]
\centering
\begin{tabular}{lc}
    \toprule
   \rotatebox[origin=c]{90}{\textbf{Predicted}} 
   & \raisebox{-.5\height}{\includegraphics[width=0.85\linewidth]{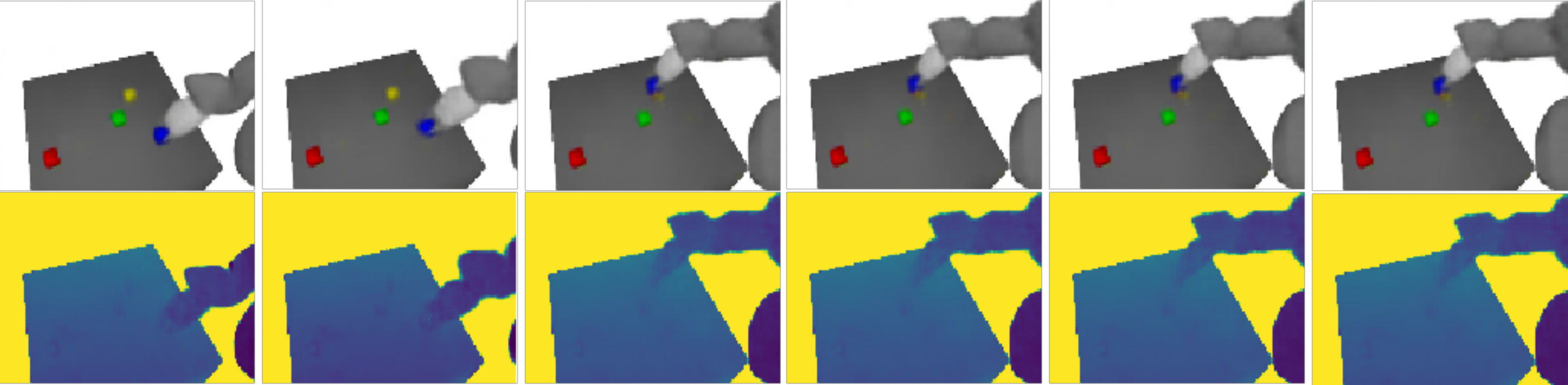}} \\
   \midrule
   \rotatebox[origin=c]{90}{\textbf{Oracle}}
   & \raisebox{-.5\height}{\includegraphics[width=0.85\linewidth]{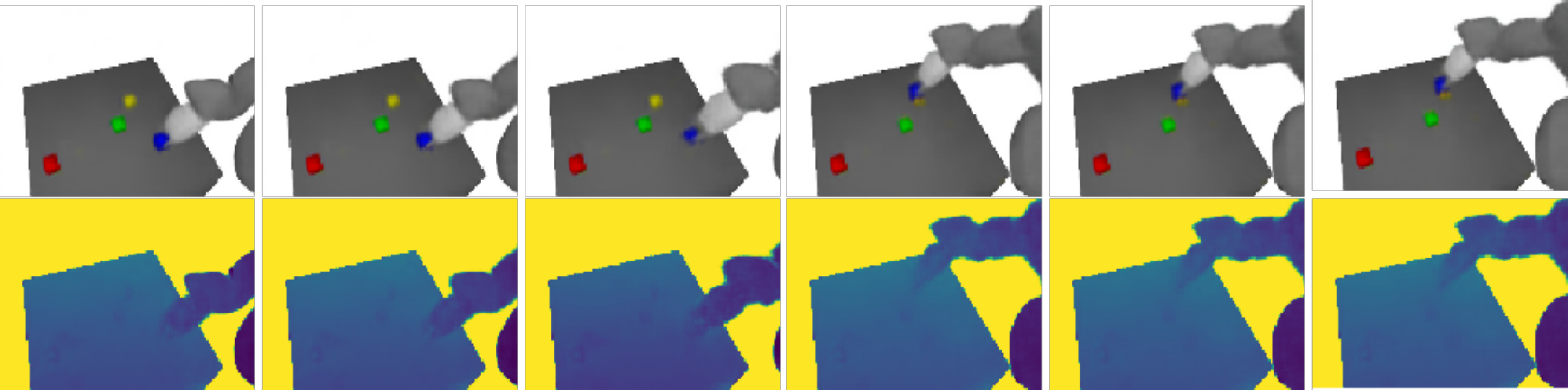}}\\
   \bottomrule
\end{tabular}
\caption{Comparison of generated subgoal predictions. Top two rows: RGB and depth images generated from predicted subgoal $G$. Bottom two rows: RGB and depth images generated from ground truth $G$ from training data.
}
\label{fig:image-comparison}
\end{figure*}

\subsection{Execution Results}

Finally, we test our model on a set of held-out scenarios, and compare to ground-truth execution. 
We compared accuracy of the estimated motion under each of three conditions: with oracle subgoals $G_t$ from the test data, with unambiguous templated language, and with natural language. Position accuracy results are shown in Table~\ref{table:extrinsic-ee-pose}.

In all cases, we compute an execution plan $\hat{G}_1,\dots,\hat{G}_5$ at the beginning and use our Predictor and Actor networks to follow this execution plan until all steps have been executed. Results are shown in Table~\ref{table:extrinsic-ee-pose}. We count successes when the block was moved to within 1.5 cm of the target in the x and y direction, and 0.5 cm z of the final position from which it was dropped.

We see only a handful of failures when the robot was sent to ground truth ``oracle'' poses, due to stochastic interactions between the objects and gripper and randomness at the control level.
$88.1$\% of execution with ground truth actions, $88.0$\% of unambiguous templated language, and $84.0$\% of natural language were successfully able to pick up a block and put it in the right area---indicating a high level of precision independent of the task specification. Grasp success rates tended to be very high. The most dramatic failures we observed were situations where one or more necessary blocks was out of the camera's viewpoint, in which case our vision-based system fails by default.

The similar performance between templated language and ground truth actions suggests that unambiguous, templated language is insufficient to demonstrate the language learning capabilities of our system.
We find our method is robust to real natural language from Mechanical Turk workers, achieving 95\% of the success rate seen on unambiguous templates.

\subsection{Subgoal Module}

\begin{table}[t]
    \centering
    \begin{tabular}{l c c c c c}
        \toprule
        Horizon & $h=1$ & $h=2$ & $h=3$ & $h=4$ & $h=5$ \\
         \midrule[1pt]
        \textbf{Template} & &&&&\\
        \hspace{1em}Verb & 84.0\% & 87.9\% & 93.0\% & 97.4\% & 100.0\% \\
        \hspace{1em}To Object & 91.5\% & 90.4\% & 93.0\% & 97.4\% & 100.0\%  \\
        \hspace{1em}With Object & 93.7\% & 91.8\% & 94.4\% & 96.8\% & 100.0\% \\
        \cmidrule{2-6}
        \hspace{1em}Overall & 82.9\% & 86.8\% & 92.5\% & 97.3\% & 100.0\%  \\
        \midrule[1pt]
        \textbf{Real Lang} & &&&&\\
        \hspace{1em}Verb & 84.2\% & 87.8\% & 92.9\% & 97.2\% & 100.0\% \\
        \hspace{1em}To Object & 87.8\% & 87.4\% & 89.8\% & 93.9\% & 98.2\% \\
        \hspace{1em}With Object & 91.4\% & 90.1\% & 92.5\% & 96.4\%  & 100.0\% \\
        \cmidrule{2-6}
        \hspace{1em}Overall & 79.6\% & 83.8\% & 89.1\% & 93.7\% & 98.2\% \\
        \bottomrule
    \end{tabular}
    \caption{Subgoal prediction accuracy ($\uparrow$) at different horizons with templated versus natural language. We see higher accuracy as we move closer to the end of the task, when the space of possible remaining plans is less ambiguous.}
    \label{table:s2s-best}
    \vskip -0.5cm
\end{table}

We analyze the language learning component of our model.
A full breakdown of subgoal prediction accuracy is given in Table~\ref{table:s2s-best}.
Performance was comparable between templated language and natural language data collected from Amazon Mechanical Turk. We see that it is more difficult to make accurate predictions on real language data.
Additionally, accuracy is remarkably consistent over time, meaning that the model properly learned the correct sequence of actions that should be executed. Accuracy farther out into the future is stable because the network knows when and how a sequence should end.

There are two major sources of error we observe in these examples. First, the difference in accuracy between Mechanical Turk language and templated language is largely explained by the ambiguity and underspecificity in Turk commands (e.g., not specifying a destination after a grasp). Second, the overall error is largely due to sequence error at transition points, where multiple possible actions are reasonable depending on whether or not the low level control has arrived at its destination. 
This further supports our hypothesis that we need to reason about all three sub-problems jointly.

\subsection{Prediction Model}

\begin{figure}[h!]
\centering
\begin{tabular}{@{}c@{\hspace{5pt}}c@{}}
\frame{\includegraphics[trim={2cm 0 0 0},clip,width=0.45\linewidth]{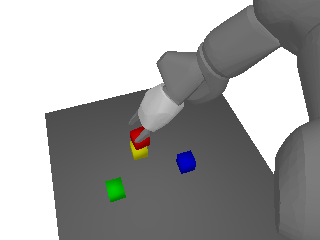}} 
& \frame{\includegraphics[trim={2cm 0 0 0},clip,width=0.45\linewidth]{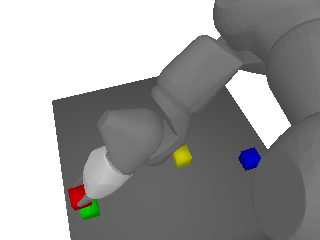}}\\[1pt]
\frame{\includegraphics[trim={2cm 0 0 0},clip,width=0.45\linewidth]{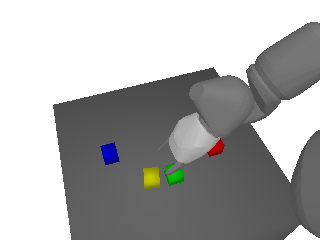}} 
& \frame{\includegraphics[trim={2cm 0 0 0},clip,width=0.45\linewidth]{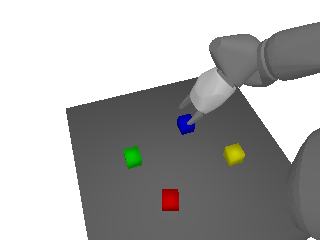}}
\end{tabular}
\caption{Results of different simulated executions. Successful grasps (top row) can be undermined by small errors in the low-level actor network that compound to create accuracy issues at execution time (bottom row).
}
\label{fig:execution-results}
\vskip -0.5cm
\end{figure}

The role of the prediction model is to generate subgoal predictions $\hat{z}_{t+1}, \dots, \hat{z}_{t+h}$ representing the $h$ next actions that the robot can take in order to perform the task.
 Fig.~\ref{fig:image-comparison} shows an example of one course of error we see during these prediction rollouts. The top two rows show a sequence of predictions coming from the sequence to sequence model, while the bottom row shows predictions using ground truth actions from our data. In this case, we see that the sequence to sequence model started the \emph{grasp} verb earlier than the ground truth execution, but both models generate good image predictions.

As we can see in Table~\ref{table:extrinsic-ee-pose}, there is persistently some error in the low-level predictions from our actor module, even when given oracle arguments Average placement error increases as we move away from the ground-truth arguments. Often failures occur because the object is not clearly visible in the first frame.

Our reconstruction results have another advantage, however, as seen in Fig.~\ref{fig:image-comparison}: they are clearly interpretable, which means that the robot can readily justify its decisions even when it does make a mistake, facilitating a human user providing a new instruction that considers this mistake. Overall, these results show that we can learn representations for a task that are sufficient for planning and execution purely from language and raw sensory data.

We performed an ablation analysis on the best prediction models to determine how much they use information from different layers. In particular, we see similar performance when training without the image loss ($89.5\%$ successful on held out test data) and without the image and object losses ($88.6\%$ successful). This suggests that our image reconstruction loss may help, and certainly does not have a negative impact.

%% file: 07_conclusions.tex
We present an approach for inferring interpretable plans from natural language and raw sensor input using prospection.
Our \dreamcell architecture predicts future world states from language and raw sensor observations, facilitating high level plan inference that can be converted into low-level execution.
Prospection enables end-to-end plan inference that is agnostic to the nature of sensory input and low-level controller modules.
In the future, using this architecture to bootstrap language understanding for execution on a real robot using sim-to-real transfer techniques could facilitate end-to-end control on a physical platform.

%% file: 08_Appendix.tex
\beginsupplement
\section{Templated vs Real Language}
Table \ref{tab:t_v_h} contains examples of the different types of language that occur when asking humans to describe the action versus using templated language. 

\begin{table}[!h]
\begin{tabular}{l l p{3.5cm} p{4.5cm}}
\multicolumn{1}{c}{\bf\large Before} &
\multicolumn{1}{c}{\bf\large After} &
\multicolumn{1}{c}{\bf\large Template} &
\multicolumn{1}{c}{\bf\large Human}\\
\toprule
\raisebox{-0.9\height}{\includegraphics[width=105pt,trim={40pt 0 0 50pt},clip]{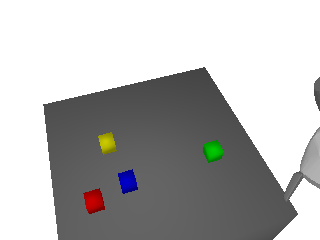}} &
\raisebox{-0.9\height}{\includegraphics[width=105pt,trim={40pt 0 0 50pt},clip]{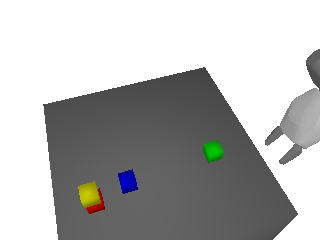}} & 
place yellow block on the red block & 
\raisebox{-0.83\height}{\makecell[{}{p{4.5cm}}]{stack warm colors\\[55pt] 
\textbf{Unknown concepts}}}
\\
\midrule
\raisebox{-0.9\height}{\includegraphics[width=105pt,trim={40pt 0 0 50pt},clip]{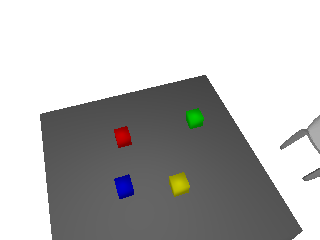}} &
\raisebox{-0.9\height}{\includegraphics[width=105pt,trim={40pt 0 0 50pt},clip]{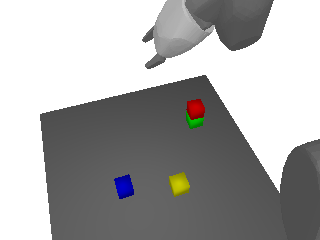}} & 
stack the red block on the green one & 
\raisebox{-0.83\height}{\makecell[{}{p{4.5cm}}]{move red right to same x and y axis as green\\[50pt] 
\textbf{Coordinate System}}}
\\
\midrule
\raisebox{-0.9\height}{\includegraphics[width=105pt,trim={40pt 0 0 50pt},clip]{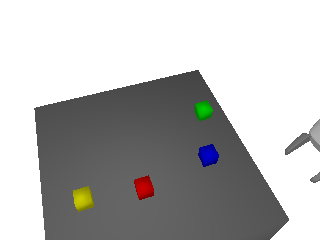}} &
\raisebox{-0.9\height}{\includegraphics[width=105pt,trim={40pt 0 0 50pt},clip]{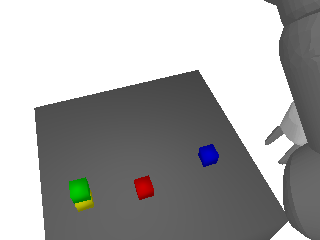}} &
place green on the yellow one & 
\raisebox{-0.83\height}{\makecell[{}{p{4.5cm}}]{move the green box forward three spaces\\[45pt] 
\textbf{Spatial language}}}
\\
\midrule
\raisebox{-0.9\height}{\includegraphics[width=105pt,trim={40pt 0 0 50pt},clip]{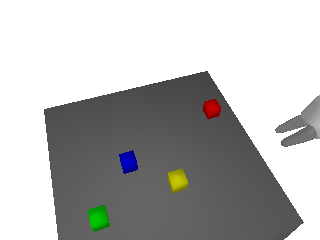}} &
\raisebox{-0.9\height}{\includegraphics[width=105pt,trim={40pt 0 0 50pt},clip]{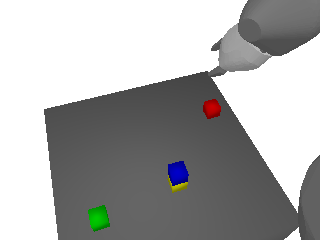}} &
stack the blue one on yellow & 
\raisebox{-0.83\height}{\makecell[{}{p{4.5cm}}]{take the blue block in your hand and raise it above the table. move the block back and to the right until it is directly above the yellow block. lower the blue block down onto the yellow block and release it\\
[10pt] \textbf{Latent details about hand movement}}}
\\
\midrule
\raisebox{-0.9\height}{\includegraphics[width=105pt,trim={40pt 0 0 50pt},clip]{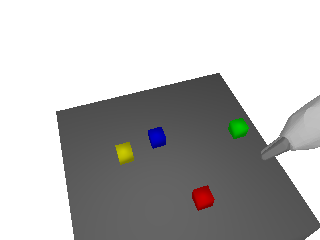}} &
\raisebox{-0.9\height}{\includegraphics[width=105pt,trim={40pt 0 0 50pt},clip]{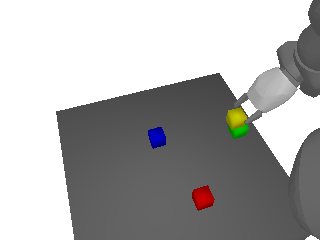}} &
put the yellow one on the green block & 
\raisebox{-0.83\height}{\makecell[{}{p{4.5cm}}]{move the yellow cube to the right until it is on top of the green cube with the front half of the yellow cube touching the far half of the top of the green cube\\[30pt] 
\textbf{Denotes specific nuance}}}
\\
\bottomrule
\end{tabular}
\caption{Above are the initial and final visual frames for each task, next to the template language and human descriptions for examples from our training set. These examples illustrate why it can be so difficult for a model to predict specific motions that correspond to a particular natural language command, and further justify our approach for visualizing robot actions before execution. Specific reasons why each description is difficult to ground are indicated in \textbf{bold}}
\label{tab:t_v_h}
\end{table}

\section{Prediction Results}

One advantage of proposed \dreamcell{} system is that it allows us to generate multiple hallucinations of possible futures. Here, we show example plans generated from four unseen test environments, given a natural-language prompt. We show predictions for the first four high level actions: \texttt{align}, \texttt{grasp}, \texttt{lift}, and \texttt{move\_to}.
Environments and trials were chosen at random, and should be indicative of performance on the prospection problem.

\begin{figure}[!t]
\centering
\begin{tabular}{@{}cc}
Prompt: & \multicolumn{1}{p{13cm}}{\large\bf ``put red on blue''}\\
\toprule
\raisebox{40pt}{1.} & \includegraphics[width=0.75\columnwidth,trim={85pt 35pt 55pt 35pt},clip]{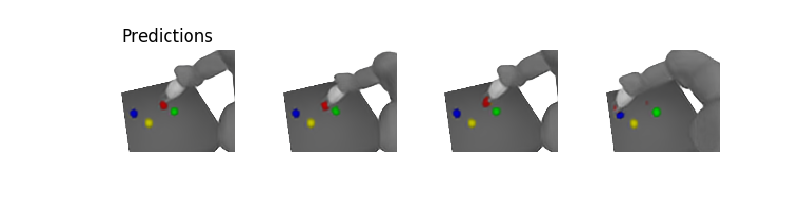}\\
\midrule
\raisebox{40pt}{2.} & \includegraphics[width=0.75\columnwidth,trim={85pt 35pt 55pt 35pt},clip]{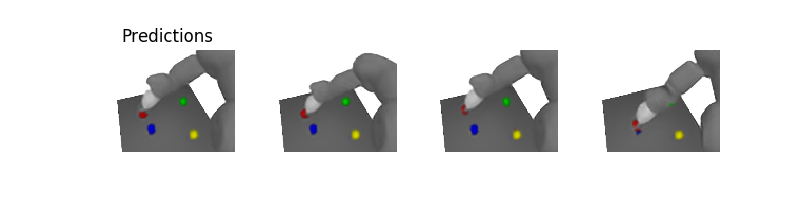}\\
\midrule
\raisebox{40pt}{3.} & \includegraphics[width=0.75\columnwidth,trim={85pt 35pt 55pt 35pt},clip]{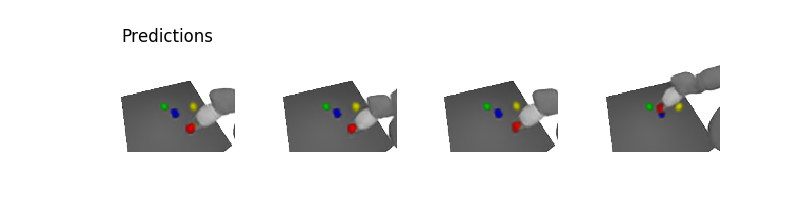}\\
\midrule
\raisebox{40pt}{4.} & \includegraphics[width=0.75\columnwidth,trim={85pt 35pt 55pt 35pt},clip]{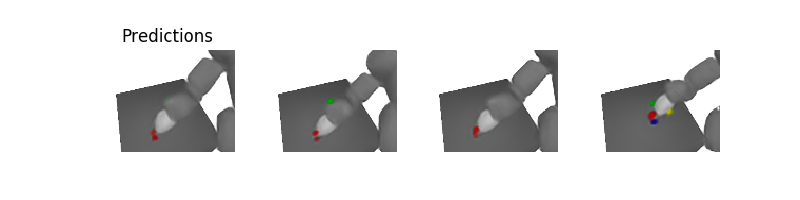}\\
\end{tabular}
\caption{Example showing predicted plans given straightforward language.}
\end{figure}

\begin{figure}[!b]
\centering
\begin{tabular}{@{}cc}
Prompt: & \multicolumn{1}{p{13cm}}{\large\bf ``put blue on the other one''}\\
\toprule
\raisebox{40pt}{1.} & \includegraphics[width=0.75\columnwidth,trim={85pt 35pt 55pt 35pt},clip]{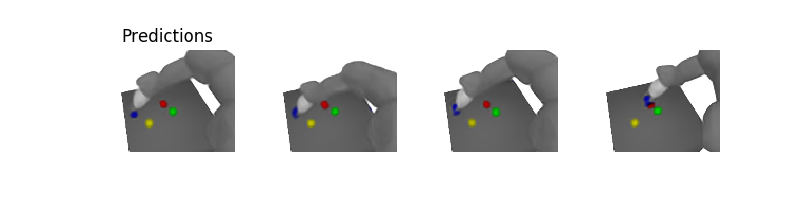}\\
\midrule
\raisebox{40pt}{2.} & \includegraphics[width=0.75\columnwidth,trim={85pt 35pt 55pt 35pt},clip]{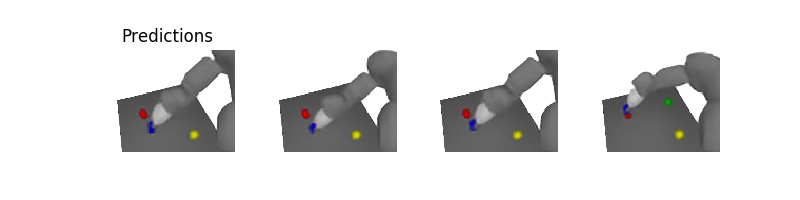}\\
\midrule
\raisebox{40pt}{3.} & \includegraphics[width=0.75\columnwidth,trim={85pt 35pt 55pt 35pt},clip]{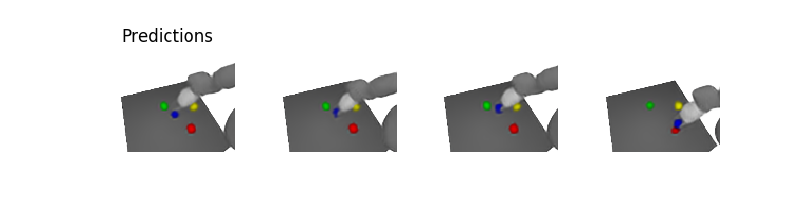}\\
\midrule
\raisebox{40pt}{4.} & \includegraphics[width=0.75\columnwidth,trim={85pt 35pt 55pt 35pt},clip]{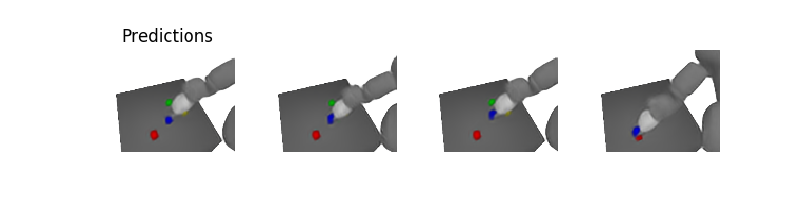}\\
\end{tabular}
\caption{Example showing predicted plans given underspecified language. The system always picks up the blue block, and correctly places it on the ``other'' one; however, it always chooses red. Handling ambiguity is left to future work.}
\end{figure}

\begin{figure}[!t]
\centering
\begin{tabular}{@{}cc}
Prompt: & \multicolumn{1}{p{13cm}}{\large\bf ``take the red block in your hand and lift it off the table and move it to the blue block and lower it and open your hand''}\\
\toprule
\raisebox{40pt}{1.} & \includegraphics[width=0.75\columnwidth,trim={85pt 35pt 55pt 35pt},clip]{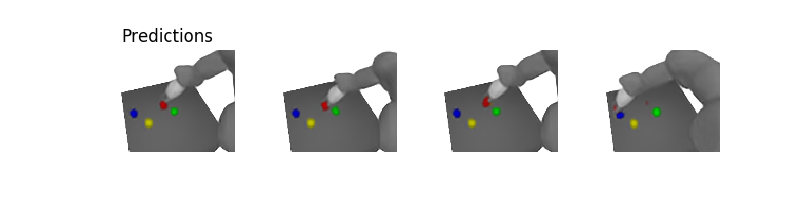}\\
\midrule
\raisebox{40pt}{2.} & \includegraphics[width=0.75\columnwidth,trim={85pt 35pt 55pt 35pt},clip]{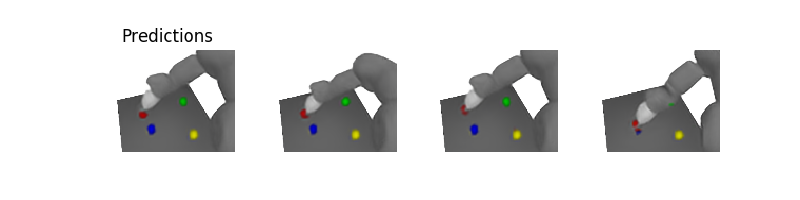}\\
\midrule
\raisebox{40pt}{3.} & \includegraphics[width=0.75\columnwidth,trim={85pt 35pt 55pt 35pt},clip]{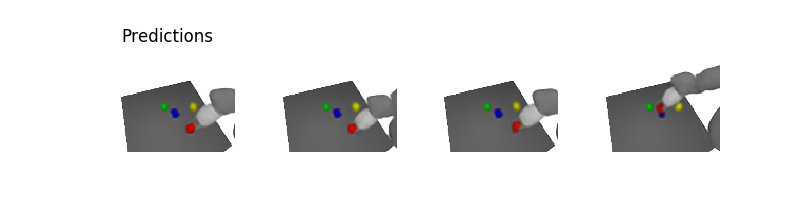}\\
\midrule
\raisebox{40pt}{4.} & \includegraphics[width=0.75\columnwidth,trim={85pt 35pt 55pt 35pt},clip]{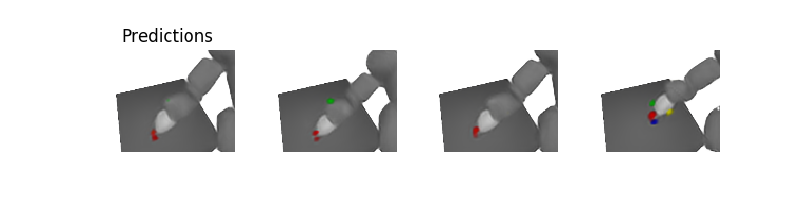}\\
\end{tabular}
\caption{Example showing predicted plans given overspecified language.}
\end{figure}

\begin{figure}[!b]
\centering
\begin{tabular}{@{}cc}
Prompt: & \multicolumn{1}{p{13cm}}{\large\bf ``stack warm colors''}\\
\toprule
\raisebox{40pt}{1.} & \includegraphics[width=0.75\columnwidth,trim={85pt 35pt 55pt 35pt},clip]{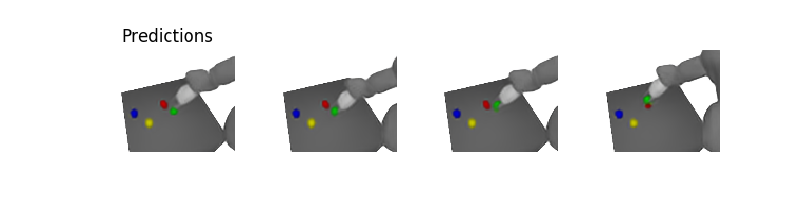}\\
\midrule
\raisebox{40pt}{2.} & \includegraphics[width=0.75\columnwidth,trim={85pt 35pt 55pt 35pt},clip]{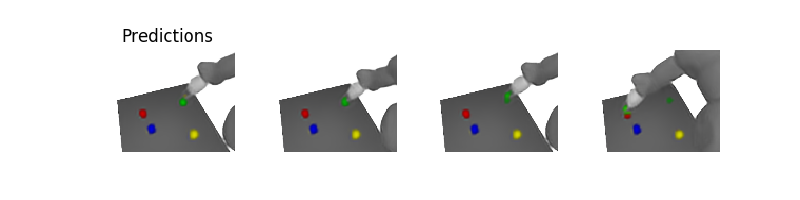}\\
\midrule
\raisebox{40pt}{3.} & \includegraphics[width=0.75\columnwidth,trim={85pt 35pt 55pt 35pt},clip]{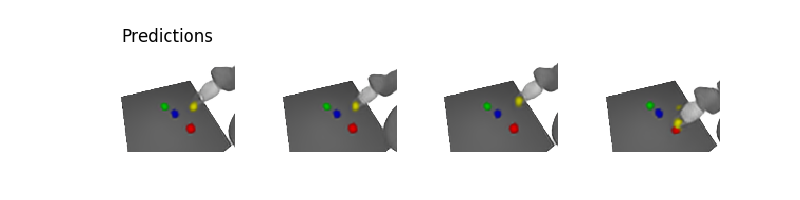}\\
\midrule
\raisebox{40pt}{4.} & \includegraphics[width=0.75\columnwidth,trim={85pt 35pt 55pt 35pt},clip]{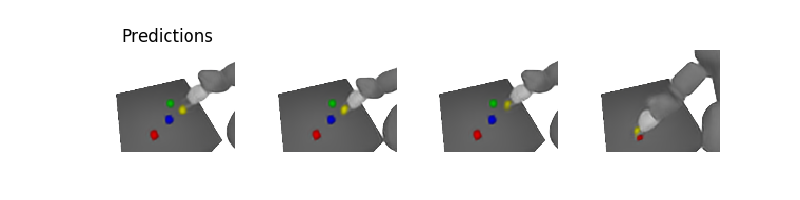}\\
\end{tabular}
\caption{Example showing predicted plans given language specified using rare ($N=1$ in train) terminology.}
\end{figure}